%% file: main.tex
\documentclass[10pt, letter, journal]{IEEEtran}

\usepackage{amsmath,amssymb,amsfonts}
\usepackage{graphicx}
\usepackage{pifont}
\usepackage{bbding}

\usepackage{amssymb}
\usepackage{amsthm}
\usepackage{xcolor}
\usepackage[figuresright]{rotating}

\usepackage{graphicx}
\graphicspath{{/}{fig/}}

\usepackage{comment}
\usepackage{array}
\usepackage{textcomp}
\usepackage{gensymb}
\usepackage{xcolor}
\usepackage{multirow}
\usepackage{booktabs}
\usepackage{enumitem}

\usepackage{mathtools}
\usepackage{breqn}
\usepackage{float}
\usepackage[noend]{algpseudocode}
\usepackage[vlined, ruled, shortend]{algorithm2e}

\usepackage{caption}
\usepackage{subcaption}

\usepackage{mathtools}
\usepackage{float}
\usepackage{hyperref}

\usepackage{pgfplots}
\pgfplotsset{compat=1.7}
\usepgfplotslibrary{groupplots}

\usepackage{multirow,tabularx}
\usepackage[ancient]{flushend}

\usepackage{makecell}

\newcommand{\RN}[1]{%
  \textup{\uppercase\expandafter{\romannumeral#1}}%
}

\usepackage{xcolor}
\colorlet{orange_maxim}{green!10!orange!90!}

\usepackage[capitalise]{cleveref}

\newlength\figureheight
\newlength\figurewidth
\usetikzlibrary{shapes.misc}

\tikzset{cross/.style={cross out, draw=black, minimum size=2*(#1-\pgflinewidth), inner sep=0pt, outer sep=0pt},
cross/.default={1pt}}

\title{
    \vspace{30pt}
    Robust 
    Multi-Modal Multi-LiDAR-Inertial Odometry and Mapping for Indoor Environments
    
}

\author{
    \IEEEauthorblockN{
        \vspace{1em}
        Li Qingqing\IEEEauthorrefmark{1}, 
        Yu Xianjia\IEEEauthorrefmark{1}, 
        Jorge Peña Queralta\IEEEauthorrefmark{1},
        Tomi Westerlund\IEEEauthorrefmark{1} \\[+0.5em]
    }
    \IEEEauthorblockA{
        \normalsize
        \IEEEauthorrefmark{1}\href{https://tiers.utu.fi}{Turku Intelligent Embedded and Robotic Systems (TIERS) Lab, University of Turku, Finland}.\\
        Emails: \textsuperscript{1}\{qingqli, xianjia.yu, jopequ, tovewe\}@utu.fi\\[+6pt] 
    }
}

\begin{document}

\maketitle
\thispagestyle{empty}
\pagestyle{empty}

\input{sec/00_Abstract.tex}
\IEEEpeerreviewmaketitle

\input{sec/01_Intro.tex}

\input{sec/02_RelatedWorks}

\input{sec/03_Methodology}

\input{sec/04_Experiments}

\input{sec/05_Conclusion}

\section*{Acknowledgment} 
This research work is supported by the Academy of Finland's AeroPolis project (Grant 348480) and the Finnish Foundation for Technology Promotion (Grants 7817 and 8089).

\bibliographystyle{IEEEtran}
\bibliography{bibliography}

\end{document}

%% file: sec/00_Abstract.tex
\begin{abstract}
    Integrating multiple LiDAR sensors can significantly enhance a robot's perception of the environment, enabling it to capture adequate measurements for simultaneous localization and mapping (SLAM). Indeed, solid-state LiDARs can bring in high resolution at a low cost to traditional spinning LiDARs in robotic applications. However, their reduced field of view (FoV) limits performance, particularly indoors. In this paper, we propose a tightly-coupled multi-modal multi-LiDAR-inertial SLAM system for surveying and mapping tasks. By taking advantage of both solid-state and spinnings LiDARs, and built-in inertial measurement units (IMU), we achieve both robust and low-drift ego-estimation as well as high-resolution maps in diverse challenging indoor environments (e.g., small, featureless rooms). First, we use spatial-temporal calibration modules to align the timestamp and calibrate extrinsic parameters between sensors. Then, we extract two groups of feature points including edge and plane points, from LiDAR data. Next, with pre-integrated IMU data, an undistortion module is applied to the LiDAR point cloud data. Finally, the undistorted point clouds are merged into one point cloud and processed with a sliding window based optimization module. From extensive experiment results, our method shows competitive performance with state-of-the-art spinning-LiDAR-only or solid-state-LiDAR-only SLAM systems in diverse environments. More results, code, and dataset can be found at \href{https://github.com/TIERS/multi-modal-loam}{https://github.com/TIERS/multi-modal-loam}.
\end{abstract}

\begin{IEEEkeywords}
    LiDAR-inertial odometry, multi-LiDAR systems, sensor fusion, solid-state LiDAR, SLAM, mapping
\end{IEEEkeywords}

%% file: sec/01_Intro.tex

\section{Introduction}
\label{sec:introduction}

Autonomous mobile robots rely heavily on spinning 3D LiDAR sensors, which offer high-quality geometric data at long ranges, with a full horizontal FoV, and robust performance across environmental conditions. As a result, this technology has found widespread application in a variety of fields, including self-driving vehicles~\cite{li2020multi}, unmanned aerial vehicles~\cite{varney2020dales}, and forest surveying systems~\cite{yang2020individual, li2020localization}, among other areas.

\begin{figure}[t]
    \centering 
    \begin{subfigure}{0.45\textwidth}
        \centering 
        \includegraphics[width=\textwidth]{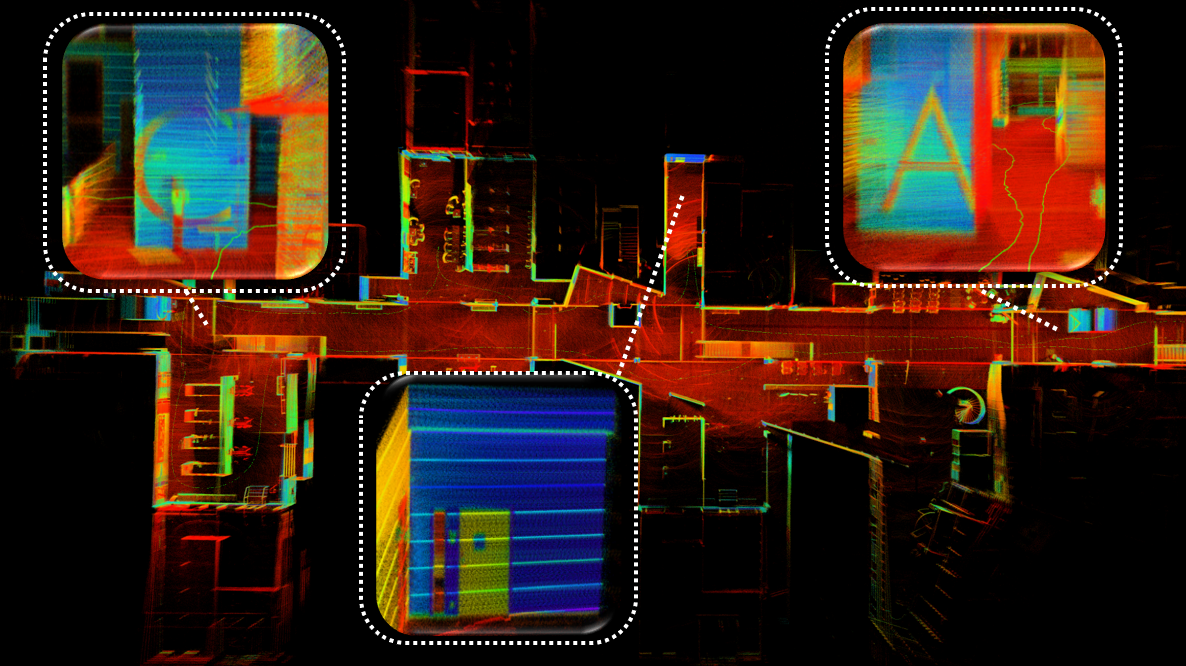} 
        \caption{Mapping result with the proposed system at a hall environment. Thanks to the high resolution of solid-state LiDAR with a non-repetitive scan pattern, the mapping result is able to show clear detail of object's surface.}  
        \label{fig:hall_mapping} 
    \end{subfigure}
    
    \vspace{0.5em}
    
    \begin{subfigure}{0.45\textwidth}
        \centering 
        \includegraphics[width=\textwidth]{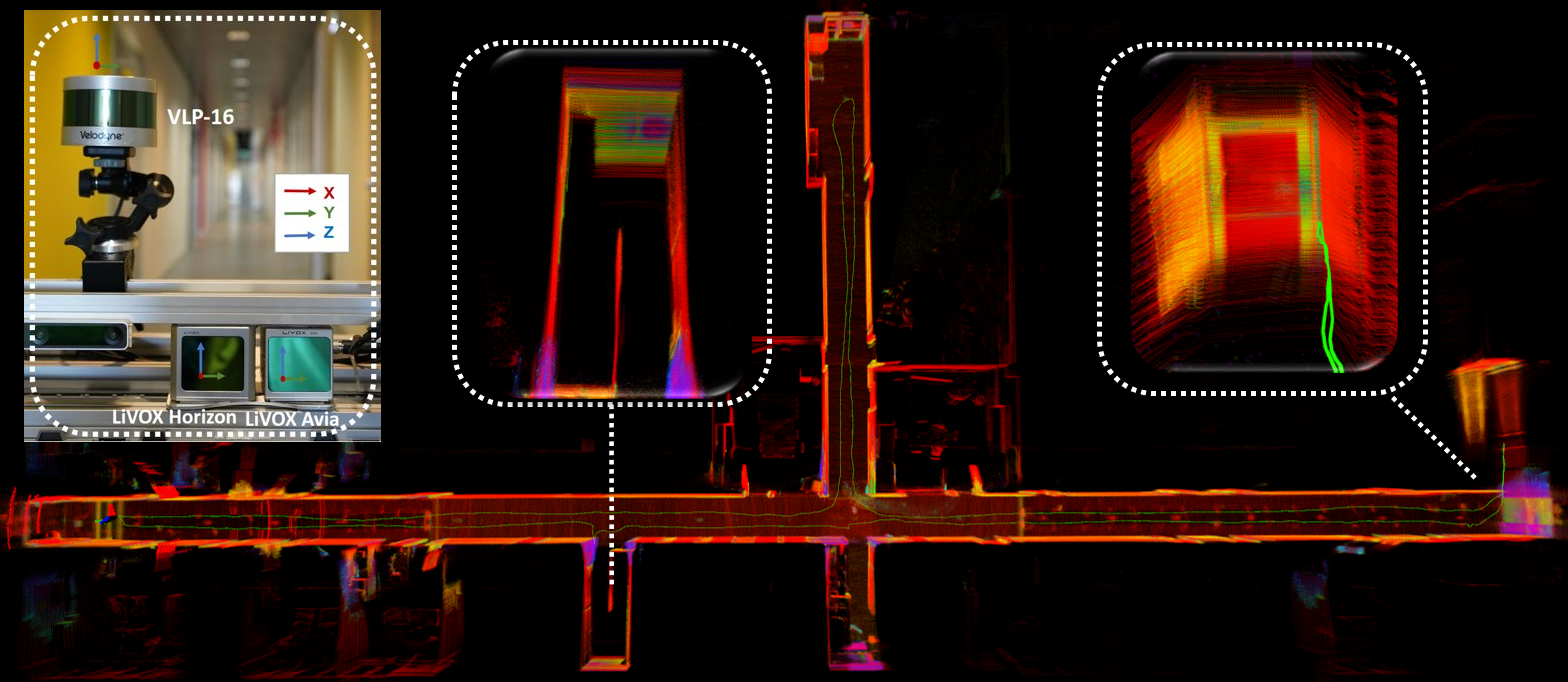} 
        \caption{Hardware and Mapping result in long corridor environment. Our proposed methods show robust performance in long corridor environments and survive in narrow spaces where 180\textdegree U-turn occurs.}  
        \label{fig:corridor_mapping} 
    \end{subfigure}
    
    \caption{Our proposed methods show high-resolution mapping results and robust performance in challenging environment}
\end{figure}

In odometry and mapping tasks, achieving denser 3D geometry measurements of the surrounding environment is crucial for enhancing 3D environment understanding. Unfortunately, spinning LiDAR with higher resolution and multiple beams can be costly due to its more complicated architecture. Low-resolution spinning LiDAR, while more affordable, produces sparse point clouds with limited features, making the problem difficult to tackle~\cite{ye2019liomap} and leading to inevitable inherent alignment errors~\cite{xu2022robust}.
  
LiDAR technology has advanced rapidly in recent years, with new sensors able of generating image-like data in addition to point clouds~\cite{tampuu2022lidar, xianjia2022analyzing}, and new solid-state LiDAR sensors that offer dense 3D point clouds with non-repetitive scan patterns, while at the same time lowering the cost~\cite{lin2020loam, li2021towards}. 
Despite the clear advantages of solid-state LiDARs, the naturally narrow horizontal FoV leads, in a similar way to monocular pinhole camera systems, to the sensing volume being blocked by objects, or even entirely occupied by a near wall, resulting in an insufficient number of feature points to estimate a 6-degree-of-freedom (6-DoF) pose. This limitation has made current solid-state-based SLAM methods challenging to apply in outdoor or large indoor scenarios~\cite{li2021towards, sier2022benchmark}. 


 Multiple studies in the literature have focused on improving LiDAR maps by integrating point clouds from multiple LiDAR sensors~\cite{jiao2021robust,chen2021backpack}. However, the low frame publishing frequency of typical LiDARs (e.g., 10\,Hz) can hinder an accurate 6-DOF pose estimation in multi-LiDAR systems. In contrast, IMUs have been widely used in state-of-the-art SLAM systems~\cite{qin2018vins, li2021towards}, due to their ability to measure acceleration and angular velocity at a high frequency (e.g., 200\,Hz) in three-dimensional space. Nonetheless, there remains a lack of methods that can effectively exploit multi-LiDAR inertial systems for odometry estimations.

To improve the robustness of the SLAM system, we propose a novel tightly-coupled multi-modal multi-LiDAR-inertial odometry and mapping system, which takes advantage of both the large horizontal FoV from a spinning LiDAR and the dense measurements from a solid-state LiDAR as Table~\ref{table:lidars_comp} shows. The proposed system first performs spatial-temporal calibration to align the timestamp and calibrate the extrinsic parameters between sensors. Then, we extract two group feature points, edge and planar points, from LiDAR data. Next, with pre-integrated IMU data, an un-distortion module is employed on Lidar point cloud data. 
Finally, the un-distorted point cloud is merged into one point cloud and sent to sliding window based optimization module. 
This work is, to the best of our knowledge, the first multi-LiDAR-inertial SLAM system able to effectively integrate LiDAR sensors with heterogeneous scan modalities within a single estimation and optimization framework. This work is inspired by the limitations found in state-of-the-art algorithms for different LiDAR sensors in our previous works~\cite{li2022dataset,sier2022benchmark}, where we show that low-cost solid-state LiDARs outperform high-resolution spinning LiDAR in an outdoor environment, while at the same time perform poorly in indoor environments. The unique characteristics and main contributions of our work can be summarized as follows: 
\begin{enumerate}
    \item Present a complete solution for multi-modal LiDAR spatio-temporal calibration and feature extraction. The method adopts an ICP-based scan-matching approach to obtain the extrinsic parameters, split-and-merge based timestamp alignment, and unified channel based feature extraction for both spinning and solid-state-lidar.
    \item Design and implementation of a novel tightly-coupled multi-modal multi-LiDAR-inertial mapping framework that is able to combine LiDARs with different scanning modalities and IMU for odometry estimations.
    \item The demonstration of a SLAM method for taking advantage of low-cost spinning LiDARs and solid-state LiDARs that outperform the state-of-the-art in high-resolution mapping with high levels of detail.
\end{enumerate}
We provide a unique open-source implementation for tightly-coupled multi-modal LiDAR and IMU fusion available to the community. Through extensive experiments, our proposed methods show state-of-the-art capabilities in various environments, with comparable odometry estimations and higher map quality. The structure of the paper is as follows. Section II surveys the existing research in multi-LiDAR systems and mapping and SLAM with solid-state LiDAR. Section III introduces our proposed mythology. Section IV delves into the details of the implementation and experimental results. Finally, Section V concludes the study and suggests future work.

%% file: sec/02_RelatedWorks.tex
\begin{table}[t]
    \centering
    \caption{Characterization of off-the-shelf LiDAR sensors based on horizontal resolution (H. Res.), vertical resolution (V. Res.), and cost.}
    \scalebox{0.95}{
    \begin{tabular}{@{}lccc@{}}  
    \toprule 
        Lidar Types       & \textbf{High H. Res.}     & \textbf{High V. Res.}  & \textbf{Low-cost}           \\[0.5ex]
     \midrule     
        Spinning, 64+ channels    & \CheckmarkBold    & \CheckmarkBold    & \XSolidBrush      \\[0.5ex]
        Spinning, 16-32 channels    & \CheckmarkBold    & \XSolidBrush      & \CheckmarkBold    \\[0.5ex] 
        Solid-State     & \XSolidBrush      & \CheckmarkBold    & \CheckmarkBold    \\[0.5ex] 
        
        Ours (solid-state + spinning-16)     & \CheckmarkBold      & \CheckmarkBold    & \CheckmarkBold    \\[0.5ex]  
    \bottomrule 
    \hspace{.42em}
    \end{tabular}  
    }
    \label{table:lidars_comp}
\end{table} 
 
\section{Related Works}
\label{sec:related_works} 

\subsection{Feature-based LiDAR odometry and mapping}

LiDAR odometry generally employs scan-matching techniques including ICP, KISS-ICP~\cite{vizzo2023kiss}, GICP~\cite{segal2009gicp}, and others to determine the relative transformation between two successive frames. Feature-based matching approaches have gained popularity as a computationally efficient alternative to full point cloud matching. For example, in~\cite{zhang2014loam}, Zhang et al. propose the registration of edge and plane features for real-time LiDAR odometry. This type of operation assumes that the LiDAR moves within a structured environment, with edge and plane points clearly identifiable from the point clouds. The matching of consecutive scans is then performed by solving a least-squares optimization problem. SLAM with features of planes for indoor environments has attracted many researchers' interests as planes ubiquitously exist in indoor environments~\cite{zhou2021plane}.

\subsection{SLAM with solid-state LiDARs}\label{sec:solid_state}  

With the development of LiDAR technology, low-cost and high-performance solid-state LiDAR has attracted significant researcher interest. The different sensing characteristics means that new challenges in point cloud registration and mapping arise. In~\cite{lin2020loam}, Lin et al. address several fundamental challenges with a robust, real-time LiDAR odometry and mapping algorithm for solid-state LiDAR (LOAM Livox) that accounts for the reduced FoV and the non-repetitive sampling patterns have been presented by taking effort on both front-end and back-end in. Other recent results have also presented tightly-coupled LiDAR-inertial odometry and mapping schemes for both solid-state and mechanical LiDARs~\cite{li2021towards, xu2021fast}. Regarding the inherently limited FoV  of a single solid-state LiDAR, a decentralized approach for simultaneous calibration, localization, and mapping utilizing multiple solid-state LiDARs was introduced in~\cite{lin2020decentralized} to enhance system resilience.

\begin{figure}[t]
    \centering  
     \includegraphics[width=0.48\textwidth]{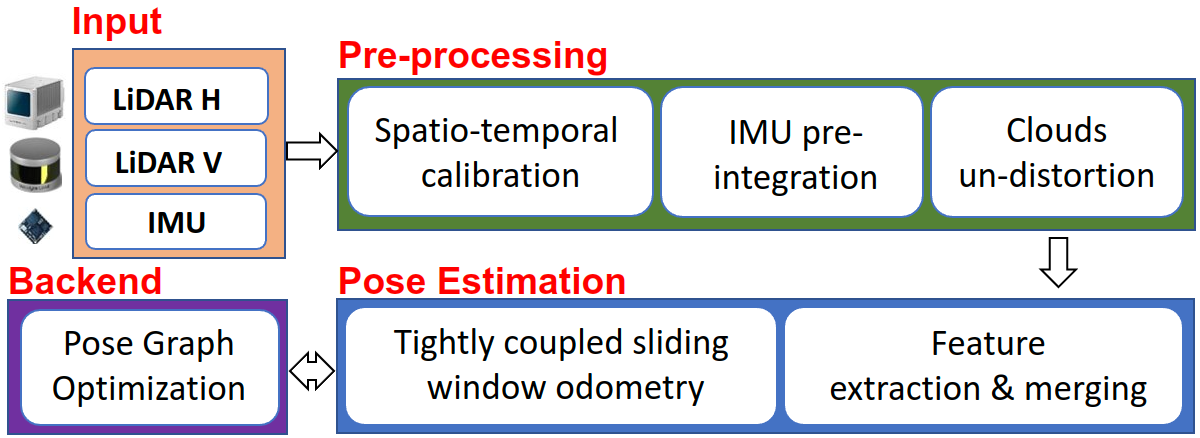}
    \caption{ The pipeline of proposed multi-modal LiDAR-inertial odometry and mapping framework.
    The system starts with preprocessing module which takes the input from sensors and performs IMU pre-integration, calibrations, and un-distortions. 
    The scan registration module extracts features and sent the features to a tightly coupled sliding window odometry. Finally, a pose graph is built to maintain global consistency. } 
    \label{fig:sys_overview} 
\end{figure} 

\begin{figure}[t]
    \centering  
    \includegraphics[width=0.48\textwidth]{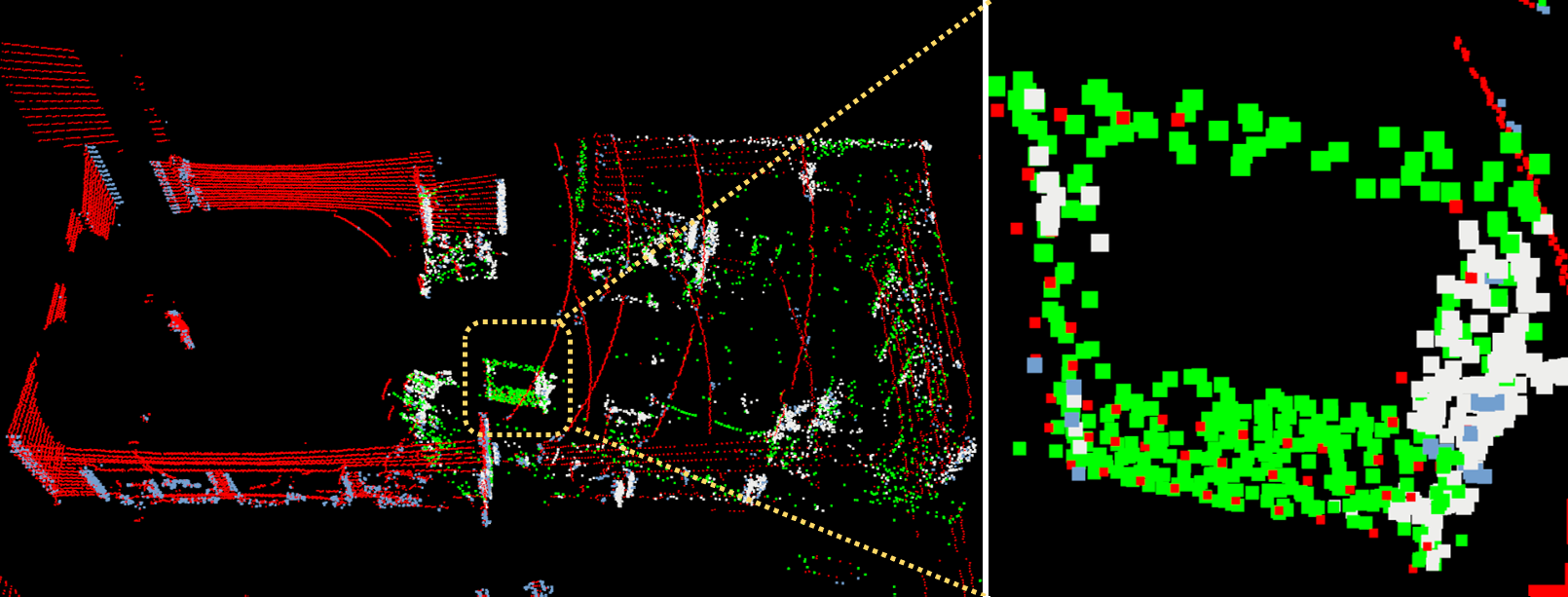}
    \caption{Extracted features points in office room environment(left) and zoom-in view of one gate object (right). Plane and edge feature points from Velodyne are in red and blue, and from Horizon are in green and white separately.}
    \label{fig:feature_pts} 
\end{figure} 

\begin{figure}[t]
    \centering  
    \includegraphics[width=0.45\textwidth]{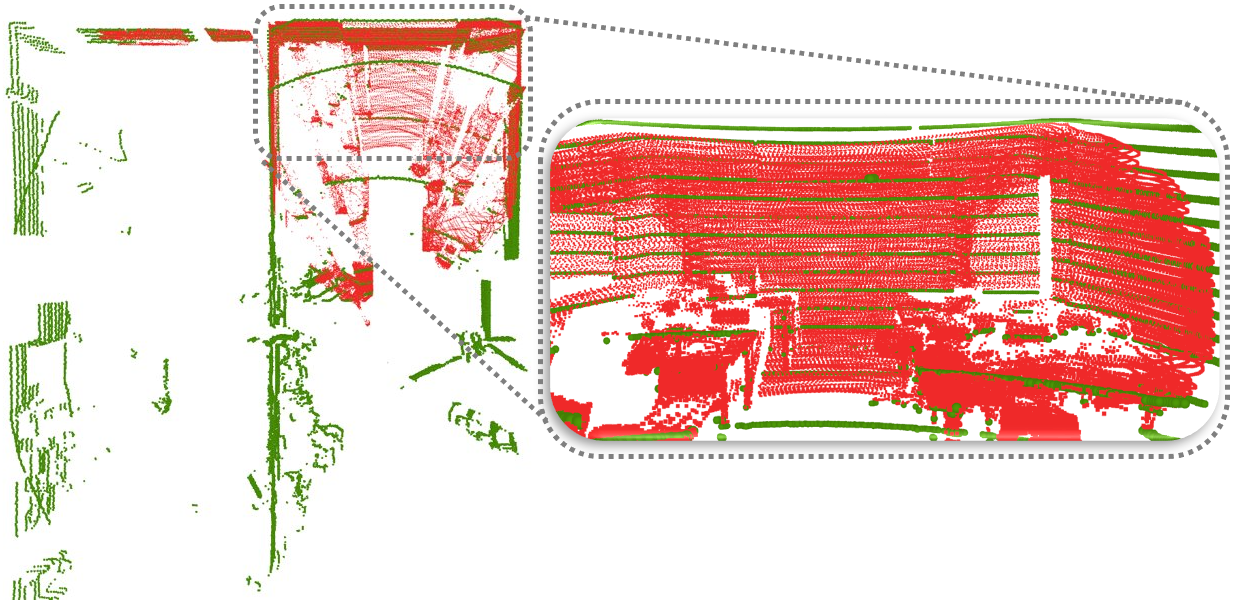}
    \caption{Multi-LiDAR extrinsic parameter calibration result in an office room environment. Red points come from Horizon and green points from VLP-16. Top view (left), and detail of the matching (right).}
    \label{fig:extrin_cali} 
\end{figure}

\subsection{Multi-modal LiDAR-based SLAM}

Relying solely on spinning LiDARs for pose estimation is suboptimal, as registering skewed point clouds or features can eventually result in substantial drift. Contemporary LiDAR SLAM systems commonly integrate data from multiple sensors to enhance accuracy. LIO-SAM was proposed as a method to eliminate the accumulated drift of LiDAR-inertial odometry over extended periods or in feature-sparse environments, by tightly coupling LiDAR and inertial measurement unit (IMU), and optionally GNSS sensors, via smoothing and mapping~\cite{shan2020lio}. SLAM approaches utilizing the fusion of LiDAR and IMU data can be found in other studies, for example, LIO-Mapping~\cite{ye2019tightly} and Fast-LIO with iterated Kalman filter~\cite{xu2021fast}. Moreover, integrating LiDAR-inertia odometry with visual-inertia odometry (VIO) can further improve accuracy while keeping the robustness either in  texture-less or feature-less environments~\cite{shan2021lvi}.

\subsection{Multi-LiDAR odometry and mapping}

More recently, the research focus has also shifted towards the fusion of data from multiple LiDARs. For example, \cite{jiao2021robust}~addresses multi-LiDAR online extrinsic calibration, odometry, and mapping, where extracted edge and planar features are utilized  and data uncertainty is modeled with Gaussian distribution. A tightly coupled LiDAR-inertial odometry and mapping approach with low drift is proposed in~\cite{nguyen2021miliom}, which utilizes features extracted from multiple time-synchronized LiDARs with complementary FoV. Rather than utilizing spinning LiDARs, a decentralized extended Kalman filter (EKF) approach for simultaneous calibration, localization, and mapping for multiple solid-state LiDARs was introduced to improve upon the limited FoV of a single solid-state LiDAR~\cite{lin2020decentralized}. However, we find a lack of work in the literature in terms of fusing multiple LiDARs with different scanning modalities, which has potential due to the different advantages they have in terms of map quality and odometry accuracy.

%% file: sec/03_Methodology.tex
\section{System Overview}

To simplify the system design, we have made follow assumptions:1)  The LiDARs are synchronized at the software level. The time offset between sensors is not considered in our system 2)The extrinsic parameters between IMU and at least one LiDAR are known. In our case, we use LiDAR's built-in IMU and use the extrinsic parameter from factory settings.

\subsection{Overview} 

The design of our system is motivated by our previous work~\cite{li2022dataset} where solid-state LiDAR shows significant performance outdoors but failed all tests indoors. To combine the high situation awareness ability and robust performance, here we proposed  multi-modal LiDAR-inertial odometry and mapping scheme. In this paper, we consider a perception system consisting of multiple modal LiDARs and IMU, and LiDAR sensors are not triggered by hardware based external clock. The pipeline of the proposed method is illustrated in Fig.~\ref{fig:sys_overview} and the hardware system is shown in  Fig.~\ref{fig:hardware_platform}. Our hardware is composed of a spinning LiDAR Velodyne VLP-16, a low-cost solid-state LiDAR Livox Horizon, and its built-in IMU.

\subsubsection{System pipeline}

The system starts with a data pre-processing module, in which IMU data are pre-integrated, extrinsic parameters between sensors are calibrated, timestamps of clouds with different starting times are aligned, the clouds from multiple LiDARs are un-distorted with IMU pre-integrated results. After pre-processing, feature point clouds representing plane and edge points are extracted and merged into one cloud. The merged feature cloud will be sent to the sliding window odometry module where the feature cloud will be matched against the local map. Together with pre-integrated IMU, fixed size of feature clouds, and local map, six-DoF egomotions and IMU parameters are estimated and optimized by keyframe-based sliding window optimization. At the backend, the system maintains a global pose graph with selected keyframes. Loop closure is detected in a keyframe basis graph using ICP, and a global graph optimization is invoked to guarantee the reconstructed map is globally consistent.

\subsubsection{Notation and Problem Formulation} 

We treat IMU coordinate as the base local coordinate indicated as $()^{b}/()^{I}$. The merged cloud will be transformed to $()^{b}/()^{I}$. We use the first keyframe received by the system as the origin of the world coordinates denoted as $()^w$. The coordinate of spinning LiDAR is denoted as $()^{v}$, and the coordinates of solid-state LiDAR are denoted as $()^{h}$. We use $\mathcal{P}_{t_1}^v = \{ \textbf{p}_1^v,\textbf{p}_2^v,...\textbf{p}_n^v\}$ be the point cloud acquired at time $t_1$ with spinning LiDAR, and $\mathcal{P}_{t_2}^h = \{ \textbf{p}_1^h,\textbf{p}_2^h,...\textbf{p}_n^h\}$ be the cloud acquired at time $t_2$ with solid state LiDAR, where $\textbf{p}_i^v$ and  $\textbf{p}_j^h$ are a point in $\mathcal{P}_{t_1}^v$ and $\mathcal{P}_{t_2}^h $.

We denote \textit{${\mathbb{F}_E}_k$} and \textit{${\mathbb{F}_P}_k$} as the edge and plane feature point cloud extracted from the LiDARs’ data at time $k$. The transformation matrix is denoted as $\textbf{T}_a^b \in SE(3)$, which transforms a point from frame $()^a$ into the frame $()^b$. $\textbf{R}_a^b\in SO(3)$ and $\textbf{t}\in R^3$ are the rotation matrix and the translation vector of $\textbf{T}_a^b$ respectively.The quaternion $\textbf{q}_a^b$ under Hamilton notation is used, which corresponds to $\textbf{R}_a^b$. $\otimes$ is used for the multiplication of two quaternions. $\textbf{q}_a^b$ and $\textbf{R}_a^b$ can be convert by Rodrigues formula. With a given point cloud from multi-modal LiDAR sensor and IMU info, the state needs to be optimized for keyframe $k$ is defined as (\ref{eq:frame_state}) where $\textbf{t}_k$ is the translation vector, ${\textbf{q}_k}$ represents orientation in quaternion, ${\textbf{v}_k}$ is the velocity, $\textbf{b}_{a_k}$ and $\textbf{b}_{g_k}$ are the bias vector of the accelerator and gyroscope. 
\begin{equation}
    {\mathbf{X}_k} = [\textbf{p}_k, \textbf{q}_k, \textbf{v}_k, \textbf{b}_{a_k}, \textbf{b}_{g_k}] \in \mathbb{R}^3 \times \mathbb{S}^3 \times \mathbb{R}^3 \times \mathbb{R}^3 \times \mathbb{R}^3 
    \label{eq:frame_state}
\end{equation}
\subsection{Pre-processing}

\subsubsection{Spatial-temporal Calibration and Initialization} 
\label{sec:st_cali}

As the extrinsic parameter between IMU and one LiDAR is known, so here we focus on calibrating the extrinsic parameters between two LiDARs. We assume the sensor platform is stationary during the extrinsic calibration process. As solid-state LiDAR with the non-repetitive pattern is able to obtain more details from the environment within the FoV, therefore, we integrated $n$ consecutive frames(e.g., ten) to new point cloud $\mathcal{P}^h_{t_1\sim t_n}$ for the calibration process. Let $\mathcal{P}^v_{t_m}$ be one cloud data obtained at $t_m \in (t_1, t_n)$ from spinning LiDAR. Generalized Iterated Closest Point (GICP)\cite{segal2009gicp} method is employed to caculate the relative transformation matrix $\textbf{T}_{v}^h$ between $\mathcal{P}^h_{t_1\sim t_n}$ and $\mathcal{P}^v_{t_m}$ from the overlapped region. The extrinsic calibration results with data collected from a classroom environment are shown in Fig. \ref{fig:extrin_cali}. As the transformation matrix $\textbf{T}_{h}^i$ between IMU and one LiDAR given by the factory, we can get $\textbf{T}_{v}^i$ by $\textbf{T}_{v}^i  = \textbf{T}_{v}^h * \textbf{T}_{h}^i$. 

We consider a system that LiDARs are not triggered with the external clock(e.g. GNSS) like~\cite{jiao2021robust}, each cloud $\mathcal{P}^h$ and $\mathcal{P}^v$ are collected at different starting timestamps. To merge clouds into one combined cloud $\mathcal{P}^m$, we need to align the starting timestamp and ending timestamp. We adopt a split-and-merge method similar to ~\cite{chen2021backpack}.The individual timestamp of $p_i^h \in \mathcal{P}^h$ and $p_i^v \in \mathcal{P}^v$ can be obtained from the sensors driver. If the timestamp for a point $p_i^v \in \mathcal{P}^v$ is not available, it also can be calculated by orientation difference~\cite{zhang2014loam}. When a new cloud $\mathcal{P}^h_{t_k}$ received at time ${t_k}$, we put all points $p_i^v \in \mathcal{P}^h_{t_k}$ to a queue $\mathbb{Q}^h$ which ordered by timestamp. When a new cloud $\mathcal{P}^v_{t_m}$ received at time $t_m$, we first get its start time $t_{m_s}$ and end time $t_{m_e}$, then all points in  $\mathbb{Q}^h$ which timestamp $t_i < t_{m_s}$will be dropped, and $t_i \in (t_{m_s}, t_{m_e})$ will be popped to a new frame $\mathcal{P}^h_{t_m}$ which share the same time domain with $\mathcal{P}^v_{t_m}$.

\subsubsection{IMU Initialization and Preintegration}
 
The IMU sensor output angular velocity and acceleration measurements are defined as $\tilde{\omega}_t$ and $\tilde{\mathbf{a}_t}$ using equations. \ref{eq:raw_ang} and \ref{eq:raw_acc}: 
\begin{equation}
    \label{eq:raw_ang}
    \boldsymbol{\omega}_t =  \boldsymbol{\omega}_t + \mathbf{b}_t^{ \boldsymbol{\omega}} + \mathbf{n}_t^{ \boldsymbol{\omega}}
\end{equation}

\begin{equation} 
    \label{eq:raw_acc}
    \tilde{\mathbf{a}_t} = \mathbf{R}_t^{WL}( \mathbf{a}_t - \mathbf{g}) + \mathbf{b_t}^{\mathbf{a}} + \mathbf{n_t}^{\mathbf{a}}
\end{equation}

Where $ \mathbf{b}_t$ is the measurement bias and $\mathbf{n}_t$ is white noise. $\mathbf{R}_t^{LW}$ is the rotation matrix from World coordinate $()^L$ to local coordinate $()^W$, $\mathbf{g}$ is the  gravity vector in world coordinate. Based on the raw measurement $\boldsymbol{\omega}_t$ and $\tilde{\mathbf{a}_t}$, we can infer the motion of the robot as follows: 
\begin{equation} 
    \label{eq:imu_pose}
    \begin{aligned} 
    \mathbf{p}_{t +  \Delta{t}} =  \mathbf{p}_t + &\mathbf{v}_t \Delta{t} + \frac{1}{2}\mathbf{g}\Delta{t}^2 +   \\
        &\frac{1}{2} \mathbf{R}_t(\mathbf{\hat{a}}_t - \mathbf{b}_t^{\mathbf{a}} - \mathbf{n}_t^{\mathbf{a}}) \Delta{t}^2
    \end{aligned}
\end{equation} 
\begin{equation}  
    \label{eq:imu_velo}
    \mathbf{v}_{t +  \Delta{t}} = \mathbf{v}_t + \mathbf{g}\Delta{t} +   \mathbf{R}_t(\mathbf{\hat{a}}_t - \mathbf{b}_t^{ \mathbf{a}} - \mathbf{n}_t^{\mathbf{a}}) \Delta{t}
\end{equation}
\begin{equation}
    \label{eq:imu_quat}
    \begin{aligned}
    \mathbf{q}_{t +  \Delta{t}} &= \mathbf{q}_t \otimes \mathbf{q}_{\Delta{t}} \\ 
    &= \mathbf{q}_t \otimes  
     \begin{bmatrix}
    exp( \frac{1}{2}  \Delta t ( \tilde{ \boldsymbol{\omega}}_t - \mathbf{b}_t^{\boldsymbol{\omega}} - \mathbf{n}_t^{\boldsymbol{\omega}}))
    \\
    1
    \end{bmatrix}
    \end{aligned}
\end{equation}

Where the  $\mathbf{p}_t, \mathbf{v}_t$ and $\mathbf{q}_t $ are the estimated position, velocity and orientation in quaternion at time $t$, $\mathbf{p}_{t +  \Delta{t}}, \mathbf{v}_{t +  \Delta{t}}$ and $ \mathbf{q}_{t +  \Delta{t}}  $ are the estimated state at time $t+\Delta{t}$.
We apply the IMU preintegration method proposed in ~\cite{qin2018vins}. The relative motion between two timestamp $\Delta{\mathbf{V}}, \Delta{\mathbf{P}}, \Delta{\mathbf{Q}}$ can be caulated based on equations \ref{eq:imu_pose}$\sim$\ref{eq:imu_quat} and will be used for initial guess in ~\ref{sec:sliding_win}.

\subsection{Multi-modal LiDAR Pose Estimation}  

\subsubsection{Union Feature Extraction}

For computing efficiency, feature extraction is essential for the SLAM system. We focus on extracting the general features that exist in different modal LiDARs and can be shared in the optimization process. 
Here we extract feature points based on~\cite{li2021towards} that selects a set of feature points from measurements according to their continuous and surface normal vector. We extend the method and adapt it to both spinning and solid-state LiDARs.
The set of extracted features consists of two subsets: plane points $\mathbb{F}_P$ and edge points $\mathbb{F}_E$. By checking the continuity, the edge feature included two types of points $\mathbb{F}_{E_l}$ and $\mathbb{F}_{E_b}$ where $\mathbb{F}_{E_l}$ represents the line feature where two surface meets, and $\mathbb{F}_{E_b}$ represents breaking points where plane end.  
 
Let $\mathcal{P}^v_t $ be the point cloud acquired at time $t$ from spinning LiDAR, $\mathcal{P}^h_t$ be the point cloud acquired from solid-state LiDAR at the same time domain after temporal alignment. If the channel number of each point $p_v \in \mathcal{P}^v_t$ is unavailable, then We first project points in $\mathcal{P}^v_t$ onto a range image based on the horizontal and vertical angle w.r.t. the origin. Each row represents data from one channel of the spinning lidar. Then the points are divided into $N$ subsets $\{\mathbf{L}^v_i\}_{i\in N}$ where $N$ is the total channel numbers of spinning lidar. For point cloud $\mathcal{P}^h_t$, we divide the point based on line number which can be obtained from Livox ROS driver~\footnote{\url{https://github.com/Livox-SDK/livox\_ros\_driver}}. Similarly, we first divided the points into $M$ subsets $\{\mathbf{L}^h_i\}_{i\in M}$ where $M$ is the total line numbers of solid-state lidar. The points in each ${\mathbf{L}^v_{i}}_{i\in N}$ and ${\mathbf{L}^h_{i}}_{i\in M}$ are ordered by timestamp. For each subset $\mathbf{L}^v_i$ in $\{\mathbf{L}^v_i\}_{i\in N}$ and $\mathbf{L}^h_j$ $\{\mathbf{L}^h_j\}_{j\in M}$, we first extract the continues points ${\mathcal{P}^v_i}_C$ and ${\mathcal{P}^h_j}_C$ by checking the depth difference with its neighbour points. If the  depth difference between the point in $\mathbf{L}^v_i$ or $\mathbf{L}^h_j$ and nearest neighbor points within the same subset is smaller than the depth threshold $d_{th}$, then the point is added to continuous points subset ${\mathcal{P}^h_i}_C$ or ${\mathcal{P}^v_j}_C$. Then we follow the feature extraction methods in~\cite{chen2021backpack}, where a scatter matrix $\boldsymbol{\Sigma}$ is calculated based on neighbor points. By analyzing the two largest eigenvalues $\lambda_1$ and  $\lambda_2$ of $\boldsymbol{\Sigma}$, the plane points are detected and labeled as a plane,  the point where two plane meet is labeled as corner features, the points where one plane ends and neighboring with discontinuous points are labeled as break points. We merge the corner points and break points together and use them as edge feature points. After this process, plane feature points ${\mathbb{F}^h_P}_{t}$ and  ${\mathbb{F}^v_P}_{t}$, edge feature points  ${\mathbb{F}^h_E}_{t}$ and ${\mathbb{F}^v_E}_{t}$ are extracted. We show extracted feature points in the office room environment in Fig.~\ref{fig:feature_pts}

\subsubsection{Feature Clouds Merging}

Instead of keeping two different system state $\textbf{X}^w_t$ for clouds $\mathcal{P}_t^h$ and $\mathcal{P}_t^v$, we merge the same type of feature point cloud into one frame and use a single system state. As the cloud $\mathcal{P}_t^h$ from solid-state lidar can be more easily blocked by near objects, with extreme cases leading to all points reflecting on a single plane (e.g., a close wall) in an indoor environment, this means that there is a certain probability of most of the points in ${\mathbb{F}^h_P}_{t}$ belonging to a single plane. This means that insufficient ${\mathbb{F}^h_E}_{t}$ points can be extracted. In this case, we treat the $\mathcal{P}_t^h$ as \textit{bad frame}, with the corresponding feature cloud not being considered to the union feature cloud ${\mathbb{F}^i_{E}}_t$ and ${\mathbb{F}^i_{P}}_t$. This ensures more consistent behavior and robust estimation across environments and over time. To detect such \textit{bad frames}, We first remove the points in ${\mathbb{F}^h_P}_{t}$ and ${\mathbb{F}^h_E}_{t}$ that is close to the origin of the sensor (e.g., 2\,m threshold), and then check the amount $n_e$ of edge point in the feature cloud ${\mathbb{F}^h_{E}}_t$. If $n_e$ is smaller than the edge feature threshold $\tau_e$ (e.g., 100 in our experiments), then the cloud $\mathcal{P}_t^h$ is treated as a \textit{bad frame}. If no such \textit{bad frame} is detected, we transform the complete feature clouds  ${\mathbb{F}^h_P}_{k}$, ${\mathbb{F}^v_P}_{t}$, ${\mathbb{F}^h_E}_{t}$ and ${\mathbb{F}^v_E}_{t}$ to the $()^i$ coordinate frame and merge them to fused, unified feature clouds ${\mathbb{F}^i_{E}}_k$ and ${\mathbb{F}^i_{P}}_k$ using the extrinsic transformation matrices $\textbf{T}_v^i$ and $\textbf{T}_h^i$, calculated as described in Section.~\ref{sec:st_cali} with Eq.~(\ref{eq:extr}).%
\begin{equation}
    \mathbb{F}^i_{E}  = \textbf{T}_v^i * \mathbb{F}_{E}^v + \textbf{T}_h^i * \mathbb{F}_{E}^h , \mathbb{F}^i_{P} = \textbf{T}_v^i * \mathbb{F}_{P}^v + \textbf{T}_h^i * \mathbb{F}_{P}^h.
    \label{eq:extr}
\end{equation}
If $\mathbb{P}_t^h$ is "bad frame", then  $ \mathbb{F}_{E}^i  = \textbf{T}_v^i * \mathbb{F}_{E}^v$ and $ \mathbb{F}_{P}^i = \textbf{T}_v^i * \mathbb{F}_{P}^v$. The union feature clouds $\mathbb{F}_{E}^i$ and $\mathbb{F}_{P}^i$ will be down-sampled before sending them into sliding window optimization module.

\subsubsection{Key\-frame Selection \& Un\-distortion}
\label{sec:sliding_win}

Given feature point cloud and preintegrated IMU within the same time domain 
${\mathbb{F}^i_E}_k, {\mathbb{F}^i_{P}}_k , {\mathbb{I}^i_{preg}}_k $ and a point cloud feature map $\mathbb{M}^w_k$ in world coordinate, the registration problem can be formulated as solving a non-linear least square problem.
%
%
The initial guess of the state $\textbf{X}^w$ is estimated with Eq.\,(\refeq{eq:guessing}):
$$ {\tilde{\mathbf{p}}}_k=  {\bar{\mathbf{p}}_{k-1}}+  {\ {\bar{\mathbf{q}}}_{k-1}}*\Delta{\mathbf{P}}_{k-1}^k$$%
\begin{equation}
{\tilde{\mathbf{q}}}_k= {\ {\bar{\mathbf{q}}}_{k-1}}*\Delta{ \mathbf{Q}}_{k-1}^k,   {\mathbf{\tilde{b_g}}_{k}} =  {\ { {\mathbf{\bar{b_g}}}}_{k-1}}
\label{eq:guessing}
\end{equation}
$$ \ {\tilde{\mathbf{v}}_k} =  {\bar{\mathbf{v}}_{k-1}}+ \ {\bar{\mathbf{q}}_{k-1}}*\Delta{\mathbf{V}}_{k-1}^k, \ {\tilde{\mathbf{b_a}}}_{k} = {\bar{\mathbf{b_a}}_{k-1}}$$
As sliding window optimization is a relatively heavy process, therefore, maintain the sparsity of the frames in the window can significantly affect the real-time performance. Here we check the IMU drift during the time interval of two consecutive keyframes, and select the frame as keyframe if the orientation difference is higher than a certain degree (e.g., 30\textdegree) or the time difference between a current frame and the last keyframe larger then certain time (e.g., 2 seconds).
Then each point in the selected keyframe will be un-distorted with $\Delta{Q}$ and $\Delta{P}$ provided by IMU pre-integration. 
Each keyframe contains deskewed feature clouds ${\mathbb{F}^i_{E}}_k$ and ${\mathbb{F}^i_{P}}_k$, pre-integrated IMU ${\mathbb{I}^i_{preg}}_k$, and initial guess of state ${\tilde{\mathbf{X} }}^w_k \sim [\tilde{\mathbf{p}}_k, \tilde{\mathbf{q}}_k, \tilde{\mathbf{v}}_k, {\tilde{\mathbf{b}_a}_k}, {\tilde{\mathbf{b}_g}_k}]$ which will be optimized by sliding window optimization.
 
\subsubsection{Sliding Window Optimization}

In this paper, we follow keyframe based tightly coupled lidar-inertial sliding window optimization strategy in~\cite{li2021towards}. The merged feature points  ${\mathbb{F}^i_{E}}_k$ and ${\mathbb{F}^i_{P}}_k$ of keyframe $k$ are treated as feature clouds that are extracted from single lidar sensor as in~\cite{li2021towards}. We build a window with $\tau$ consecutive keyframes where the states that need to be optimized for each frame are $\tilde{\textbf{X}^w} = [\tilde{\textbf{X} }^w_1,\tilde{\textbf{X} }^w_2,...,\tilde{\textbf{X} }^w_{\tau}]$. The optimal state can be obtained by minimizing the function:
\begin{equation}
     \min_{ \tilde{\textbf{X}}} \{  || \mathbb{D}_{prior}(\tilde{\textbf{X}^w})||^2  \\ +  \sum_{k=1}^{\tau} \mathbb{D}_{L}(\tilde{\textbf{X}^w_k}) +  \sum_{k=1}^{\tau} \mathbb{D}_{I}(\tilde{\textbf{X}^w_k}) \} 
      \label{eq:slide_error}
\end{equation}

Where $||\mathbb{D}_{prior}(\tilde{\textbf{X}^w})||^2 $ represents the prior residual term which is generated by marginalizing oldest frames before the current window via Schur-complement~\cite{qin2018vins},$\mathbb{D}_{I}(\tilde{X^w})$ represents the pre-integrated IMU terms as defined in~\cite{li2021towards}.$ \mathbb{D}_{L}(\tilde{\textbf{X}^w_k})$ is lidar term defined as (\ref{eq:lidar_term}).
\begin{equation}
    \sum_{a=1}^{m}(\mathbb{D}_e(\textbf{X}^w_k, \textbf{p}^i_{k,a}, \mathbb{M}^w_k))^2 + \sum_{b=1}^{n}  (\mathbb{D}_s(\textbf{X}^w_k, \textbf{p}^i_{k,b}, \mathbb{M}^w_k))^2
    \label{eq:lidar_term}
\end{equation}

$\mathbb{D}_e$ is the point-to-edge  residual term defined as (\ref{eq:edge_cost}) and $\mathbb{D}_s$ point-to-plane residual term defined as (\ref{eq:plane_cost}). 

\begin{equation}
    \mathbb{D}_e( \textbf{X}^w,\textbf{p}^i,\mathbb{M}^w)) = \frac{||( \textbf{p}^w - 
 \boldsymbol{\acute{e}^w}  )\times( \textbf{p}^w -  \boldsymbol{\grave{e}^w}   ) ||}{ || \boldsymbol{\acute{e}^w} -  \boldsymbol{\grave{e}^w}||}
 \label{eq:edge_cost}
\end{equation}

\begin{equation}
    \mathbb{D}_s( \textbf{X}^w,\textbf{p}^i,\mathbb{M}^w)) =  | \textbf{n}_s^T\textbf{p}^w + 1/|| \textbf{n}_s |||  
    \label{eq:plane_cost}
\end{equation}

where $\textbf{p}^i$ represents a feature point belonging to ${\mathbb{F}_{E}}_k, {\mathbb{F}_{P}}_k$. Then, $\textbf{p}^w = \textbf{R(q)}\textbf{p}^i + \textbf{t}$ represents the scan point $\textbf{p}^i$ at local frame $()^i$, which is transformed to world frame $()^w$ given the state estimation $[\textbf{q}, \textbf{t}]$ in $\textbf{X}^w$. We denote by $\boldsymbol{\grave{e}^w}$ and $\boldsymbol{\acute{e}^w}$ the two closest corresponding edge feature points on the feature map $\mathbb{M}^w$, while $\textbf{n}^w_s$ is the plane normal vector that is calculated by neighbor plane feature points in the $\mathbb{M}^w$ cloud. We solve the non-linear Eq.~(\ref{eq:slide_error}) using the Ceres Solver toolbox~\cite{ceressolver}. To ensure global consistency, we also maintain a pose-graph structure with optimized states $\textbf{X}^w$ and pre-integrated IMU measurements as optimization constraints.

%% file: sec/04_Experiments.tex
 \begin{figure}[t]
    \centering  
    \includegraphics[width=0.46\textwidth]{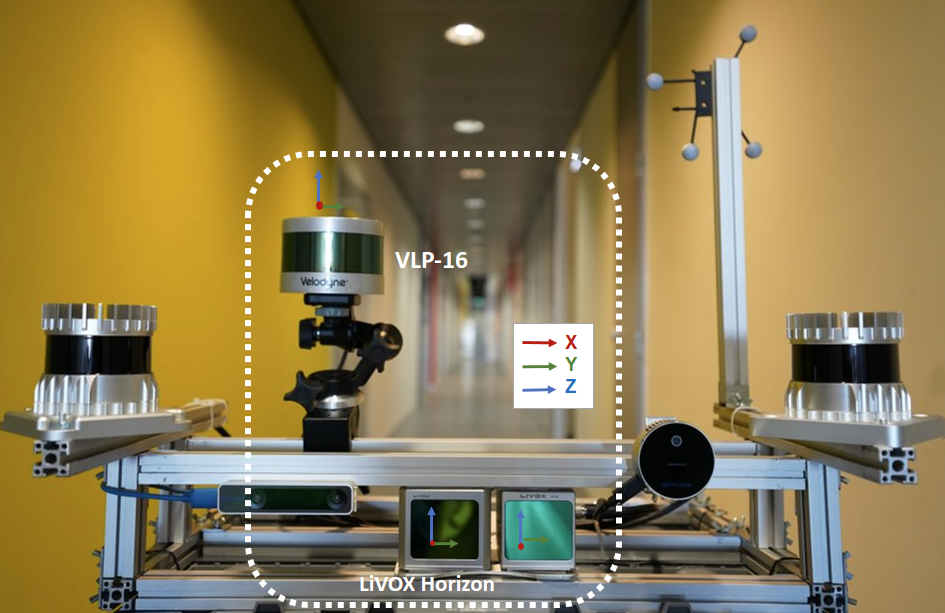}
    \caption{Hardware platform used for data acquisition. The sensors used in this work are the Livox Horizon LiDAR, with its built-in IMU, and the Velodyne VLP-16 LiDAR. The platform is mounted on a moving ground vehicle.}
    \label{fig:hardware_platform} 
\end{figure}

 \begin{figure*}
    \hfill
    \centering 
    \begin{subfigure}[b]{0.48\textwidth}
        \centering
        \includegraphics[width=\textwidth]{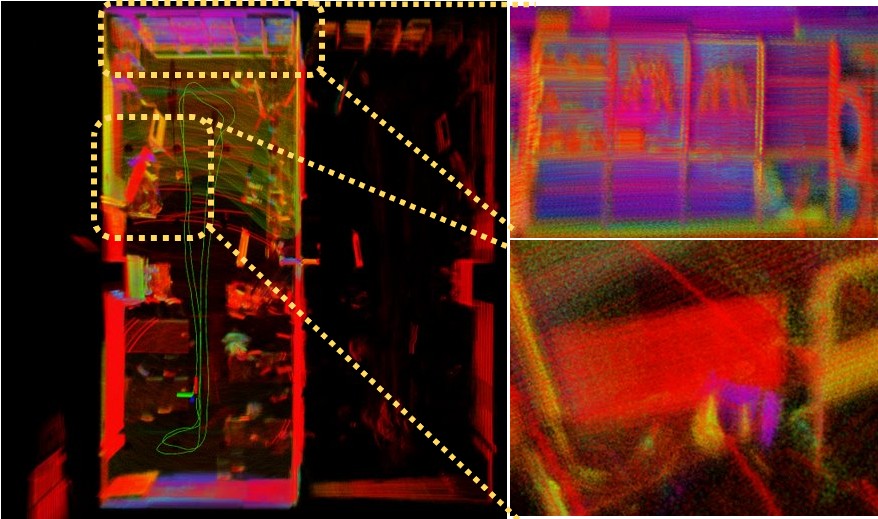}
        \caption{Map results generated by our methods in HVI mode.}
        \label{fig:map_details_a}
    \end{subfigure}
    \begin{subfigure}[b]{0.48\textwidth}
         \centering
         \includegraphics[width=\textwidth]{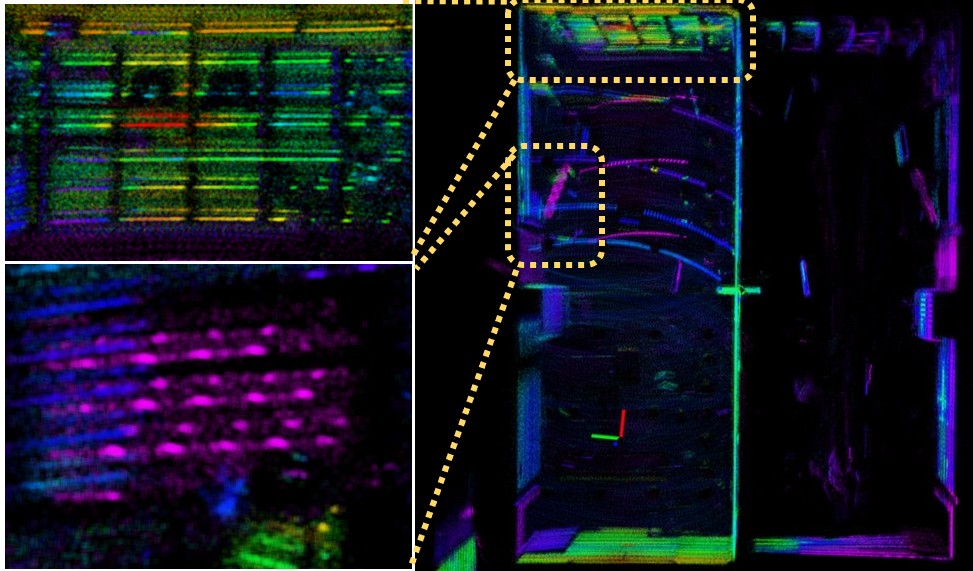}
         \caption{Map results generated Fast\_LIO in VI mode.}
         \label{fig:map_details_b}
     \end{subfigure} 
     \caption{Qualitative comparison of map details in the office room dataset sequence. The color of the points represents the reflectivity provided by raw sensor data. The point size is 1\,$cm^3$, and transparency is set to 0.05. The middle two columns show the zoom-in view of the wall (top) and TV (bottom).}
     \label{fig:map_comp}
     \hfill
\end{figure*} 
\section{Evaluation}\label{sec:evaluation}

\subsection{Sensor Configuration and Implementation} 
We implement the proposed multi-modal multi-LiDAR-inertial odometry and mapping system in C++ with ROS melodic environment that can be shared in the robotic community. The system shown in Fig.~\ref{fig:sys_overview} is structured in four nodes: preprocessing, feature extraction, scan registration, and graph optimization. The factor graph optimization is maintained by GTSAM 4.0~\cite{dellaert2012factor}, and non-linear optimization is performed by Ceres Solver 2.0~\cite{ceressolver}.
The framework proposed in this paper is validated using datasets gathered by Velodyne VLP-16 (V), Livox Horizon (H) 3D LiDAR, and its built-in IMU (I). The VLP-16 measurement range is up to 100\,m with an accuracy of \textpm\,3cm. It has a vertical FoV of 30\textdegree(\textpm\,15\textdegree) and a horizontal FoV of 360\textdegree. The 16-channel sensor provides a vertical angular resolution of 2\textdegree and The horizontal angular resolution varies from 0.1\textdegree to 0.4\textdegree. For solid-state LiDAR, we selected Livox Horizon, which is designed  with an FoV of 81.7\textdegree\,\texttimes\,25.1\textdegree. Horizon was scanning at 10\,HZ and reaches a similar but more uniform FoV coverage compared with typical 64-line mechanical LiDARs.  
The extrinsic parameter between Horizon and its built-in IMU is provided by factory instruction. The sensors are connected to a laptop directly with Ethernet and synchronized with software-based precise timestamp protocol (PTP)~\cite{lixia2012software}. We run ROS drivers for Velodyne and Horizon and recorded the data in rosbag format.

\subsection{Qualitative Experiment}
 
\begin{figure}[t]
    \centering  
     \includegraphics[width=0.5\textwidth]{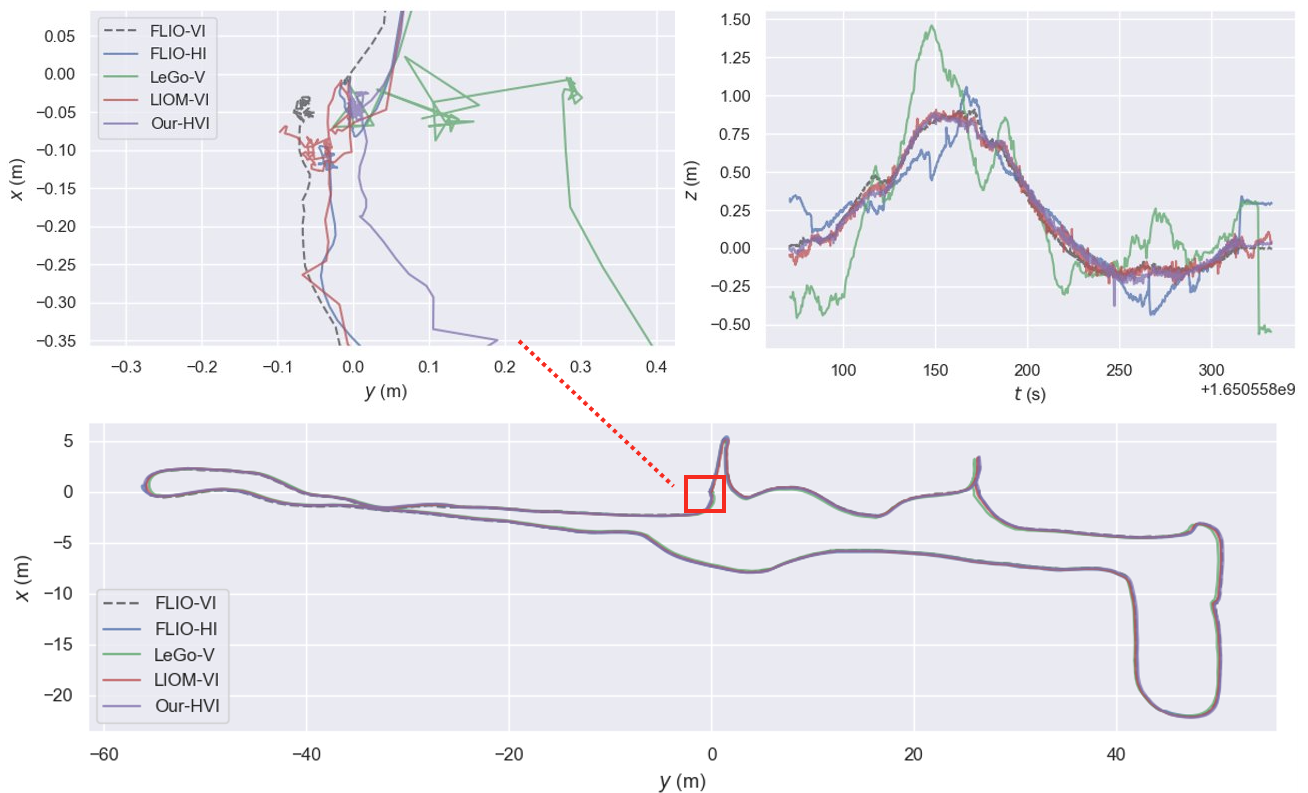}
    \caption{The trajectory result on dataset Hall. Our proposed methods show the smallest error when returning to the start point. The trajectory from different methods (bottom), the zoom-in view of starting and ending point(top left), the changes along Z-axis(top right)}
    \label{fig:hall_traj} 
\end{figure}

 \begin{table}[t]
    \centering
    \caption{End-to-end position error in meters (N/A when odometry estimations diverge; V: Velodyne VLP-16, H: Livox Horizon, I: IMU). Numbers in bold indicate the best performance, while underscored numbers indicate the second best in each environment.} 
    \begin{tabular}{@{}lccc@{}}
        \toprule 
            Dataset         & Hall   & Corridor  & Office                 \\[0.5ex]
         \midrule
            LeGo (V)    & 0.567   & 0.336   &   0.127                     \\  [0.5ex] 
            FLIO (HI / VI)      & 0.109 / \underline{0.069}  & N/A / \textbf{0.062}  & N/A / 0.188                \\  [0.5ex]  
            LIOM (HI / VI)  & N/A / 0.736     & N/A / 1.951      & NA /  \textbf{0.102}        \\  [0.5ex]  
            Ours (HI / VI)  & N/A / 0.107     & N/A / 0.132      & NA / 0.165                 \\  [0.5ex]  
            Ours (HVI)   & \textbf{0.051}  & \underline{0.085}      &  \underline{0.124}         \\  [0.5ex]   
        \bottomrule 
        \hspace{.42em}
    \end{tabular}    
    \label{table:pose_error}
\end{table} 

\begin{figure}[t]
    \centering  
     \includegraphics[width=0.5\textwidth]{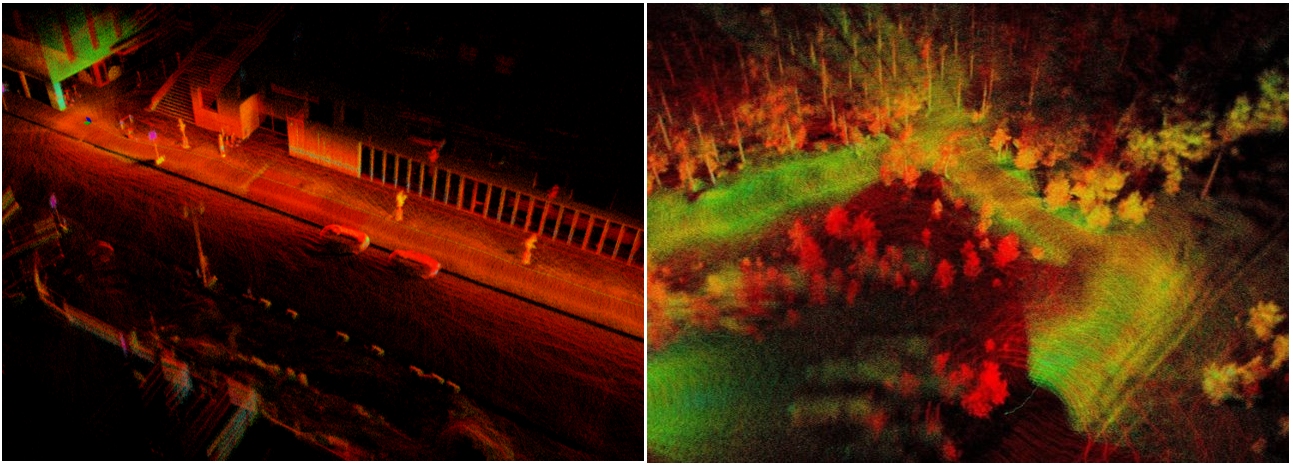}
    \caption{Mapping results with the proposed method outdoors: urban street (left) and forest environment (right).}
    \label{fig:outdoor_test} 
\end{figure}

From our previous research~\cite{li2022dataset, sier2022benchmark}, a tightly coupled solid-state LiDAR-inertial system shows competitive performance outdoors but poorly in indoor environments. Therefore, here we aim to compare our proposed system with a typical and challenging indoor environment: an office room, a long corridor, and a large hall. The data are gathered with the platform as Fig.~\ref{fig:hardware_platform} shows at ICT-City in Turku, Finland.

We compare our proposed method with several state-of-the-art SLAM algorithms: LeGO-LOAM~\cite{shan2018lego}~\footnote{\url{https://github.com/RobustFieldAutonomyLab/LeGO-LOAM}}, Fast-LIO~\cite{xu2021fast}~\footnote{\url{https://github.com/hku-mars/FAST_LIO}} and LILI-OM~\cite{li2021towards}~\footnote{\url{https://github.com/KIT-ISAS/lili-om}}. LeGO-LOAM is LiDAR only odometry, Fast-LIO and LILI-OM are tightly coupled LiDAR inertial odometry which can both work with solid state LiDAR and spinning LiDARs. Fast-LIO features with  a tightly-coupled iterated extended Kalman filter framework and iKD-tree data structure which show efficient and robust performance~\cite{xu2021fast}. Similar to our proposed method, LILI-OM employs keyframe-based sliding window optimization but only fuse single LiDAR and  pre-integrated IMU measurements. As Fast-LIO, LI
During the experiments, we use the default configurations from the official Github repository, and loop closure detection is off for each method.
%
To compare the odometry accuracy, all three datasets were starting and ended at the same place. The mean square distance (MSE) between the starting and ending positions is treated as the error. The results generated by selected methods in all dataset shows in Table~\ref{table:pose_error}. 

\subsubsection{Hall}

This data was recorded at a hall environment around 127m $\times$ 35 m to compare odometry and mapping performance in a relatively large indoor environment. The recording started at a narrow space and was followed by a 180\textdegree U-turn where most of the FoV of the solid-state-LiDAR are covered by near walls, therefore, solid-state LiDAR only based methods cannot receive enough features that might cause huge drift. 
We show the trajectory in Fig.~\ref{fig:hall_traj} and mapping results in Fig.~\ref{fig:hall_mapping}. From the position error shown in Table~\ref{table:pose_error}, our proposed methods show the best performance reaching 5.1\,cm, and Fast-LIO (VI) shows the second best performance.

\subsubsection{Corridor}

The corridor environment is another challenging environment as low-resolution LiDARs might not get enough feature points from the environment to perform robust localization. The data sequence was recorded at a 60\,m long corridor. The mapping results are shown in Fig.~\ref{fig:corridor_mapping}. Fast-LIO (VI) performs the best, while our proposed method follows closely in performance.
 
\subsubsection{Office}

Another data sequence is recorded in a small office room with a size of $12\,m \times 3.7\,m$. To make the mapping task more challenging in this environment, we perform several fast 180\textdegree U-turns during the data recording. From the result, we can see LIOM (VI) shows the best performance while our proposed method with HVI ranks second.

\subsection{Outdoor Mapping}

For the sake of completeness, we also test the proposed method with an outdoor data sequence on a city road and a forest environment. The resulting map is shown in Fig.~\ref{fig:outdoor_test}. A qualitative analysis of the map point cloud shows a high level of detail. However, the extrinsic calibration method has been designed for indoor environments with enough edge and planar features within the overlapped FoV between the two LiDAR sensors; therefore, the extrinsic parameters are not well calibrated in the forest environment, which leads to a decrease in sharpness in the final map. In any case, this demonstrates the potential for generalization to more unstructured environments and enables high-density mapping even with sub-optimal extrinsic parameters calibration

Our proposed multi-modal multi-LiDAR-inertial method demonstrates competitive and consistent performance compared to other selected methods. However, we observed that our method did not always outperform other approaches, despite utilizing more measurements from the environment. One possible reason for this could be inaccurate time synchronization between the sensors, 
our time synchronization between LiDARs is on the software level with sub-microsecond accuracy. The inaccurate timestamp for each point will bring the error to the system during the cloud undistortion and cloud merging steps, which is hard to eliminate. 

From the results, the VI-based method is able to track the position of the sensor in all dataset sequences. However, the HI based methods fail in most of the sequences except for Fast\_LIO(VI) in the large hall environment. To understand the difference between each LiDAR, we also test our methods with Velodyne-Inertial odometry. The results show the accuracy of the proposed method VI is less accurate than the HVI which indicates the horizon LiDAR can improve the system's performance. It is also worth noting that our focus is on integrating various multi-modal LiDAR sensors at the point cloud registration and feature extraction stages while aiming at a more consistent framework across environments.

 \subsection{Mapping Quality Comparison}
 
 One of the key benefits of the multi-LiDAR system is its high perception awareness ability. Here we compare the mapping quality in terms of resolution. Part of the mapping results by our proposed methods has shown in Figure.~\ref{fig:hall_mapping} and~\ref{fig:corridor_mapping} where the color represents intensity value. From the result, we can see many objects (e.g., door, wall letters).
 We compare the mapping result between our proposed method in Fig.~\ref{fig:map_details_a} and Fast\_LIO (VI) in Fig.~\ref{fig:map_details_b}. Our method shows the most stable performance in our experiments in an office room environment. By zooming in the same area, we can see a more uniform point cloud from the wall and a TV in the map generated with our method.

\subsection{Runtime Analysis}

Our evaluations were conducted on a laptop with an Intel Core i7-10875H CPU and 64\,GB RAM on Unbuntu 18.04.6 LTS system. 
We show the average runtime per frame in Table~\ref{tab:runtime} and feature numbers in Table~\ref{tab:feature_num}. Preprocessing and feature extraction are lightweight. Runtime is dominated by the sliding-window-based pose estimation and optimization.

\begin{table}[t]
    \centering
    \caption{Analysis of processing time (ms) for the different algorithm stages on an Intel i7-10875H CPU.}
    \label{tab:runtime}
    \begin{tabular}{@{}lccc@{}} 
        \toprule 
        \textbf{ }       & \textbf{Hall} & \textbf{Corridor} &  \textbf{Office} \\  
        \midrule 
        Pre-processing stage        & 4.14         & 4.44            & 4.37 \\  
        Multi-LiDAR feature extraction  & 80.32        & 86.74          & 94.21 \\   
        Pose estimation, optimization     & 139.71       & 123.51         & 132.73\\  
        \bottomrule
        \hspace{.42em}
    \end{tabular}
\end{table}

\begin{table}[t]
    \centering
    \caption{Average number of feature points per frame extracted from the different LiDARs in the three tested environments. Only points in a range between 2\,m, and 50\,m are considered.}
    \label{tab:feature_num}
    \begin{tabular}{@{}lcccccc@{}} 
        \toprule 
        \textbf{ }       &\textbf{V-raw} & \textbf{V-edge} &\textbf{V-plane} &\textbf{H-raw} & \textbf{H-edge} &\textbf{H-plane} \\  
        \midrule 
        \textbf{Hall}    & 17370       & 201   & 15680   & 21109  & 405  & 1934\\  
        \textbf{Corridor}& 5965        & 101   & 5493    & 14623  & 302  & 4007 \\   
        \textbf{Office}  & 20235       & 329   & 18430   & 17462  & 409  & 1029\\  
        \bottomrule
        \hspace{.42em}
    \end{tabular}
\end{table}

%% file: sec/05_Conclusion.tex
\section{Conclusion}\label{sec:conclusions} 

We have presented in this paper a tightly coupled multi-modal multi-LiDAR-inertial odometry and mapping framework with sliding window optimization for pose estimation. This is, to the best of our knowledge, the first SLAM algorithm to leverage the advantages of both spinning LiDARs and solid-state LiDARs within a single framework. Specifically, we have focused on demonstrating that high-robustness odometry and high-quality mapping are possible with an adequate combination of low-cost sensors. The proposed system effectively fuses the high situational awareness through dense point clouds in a solid-state LiDAR with the larger FoV from a spinning LiDAR. Despite the odometry accuracy being compared to other methods in certain environments, key properties of our method are consistency across environments and, specially, higher-quality maps where environment details can be appreciated.

%
%
In the next steps, we plan to explore further multi-modal LiDAR sensors providing image-like data and multi-IMU fusion within a single framework. We expect to focus further on online calibration in such more heterogeneous systems.


%% file: main.bbl
\begin{thebibliography}{10}
\providecommand{\url}[1]{#1}
\csname url@samestyle\endcsname
\providecommand{\newblock}{\relax}
\providecommand{\bibinfo}[2]{#2}
\providecommand{\BIBentrySTDinterwordspacing}{\spaceskip=0pt\relax}
\providecommand{\BIBentryALTinterwordstretchfactor}{4}
\providecommand{\BIBentryALTinterwordspacing}{\spaceskip=\fontdimen2\font plus
\BIBentryALTinterwordstretchfactor\fontdimen3\font minus
  \fontdimen4\font\relax}
\providecommand{\BIBforeignlanguage}[2]{{%
\expandafter\ifx\csname l@#1\endcsname\relax
\typeout{** WARNING: IEEEtran.bst: No hyphenation pattern has been}%
\typeout{** loaded for the language `#1'. Using the pattern for}%
\typeout{** the default language instead.}%
\else
\language=\csname l@#1\endcsname
\fi
#2}}
\providecommand{\BIBdecl}{\relax}
\BIBdecl

\bibitem{li2020multi}
Q.~Li, J.~P. Queralta, T.~N. Gia, Z.~Zou, and T.~Westerlund, ``Multi-sensor
  fusion for navigation and mapping in autonomous vehicles: Accurate
  localization in urban environments,'' \emph{Unmanned Systems}, vol.~8,
  no.~03, 2020.

\bibitem{varney2020dales}
N.~Varney, V.~K. Asari, and Q.~Graehling, ``Dales: a large-scale aerial lidar
  data set for semantic segmentation,'' in \emph{Proceedings of the IEEE/CVF
  Conference on Computer Vision and Pattern Recognition Workshops}, 2020.

\bibitem{yang2020individual}
J.~Yang, Z.~Kang, S.~Cheng, Z.~Yang, and P.~H. Akwensi, ``An individual tree
  segmentation method based on watershed algorithm and three-dimensional
  spatial distribution analysis from airborne lidar point clouds,'' \emph{IEEE
  Journal of Selected Topics in Applied Earth Observations and Remote Sensing},
  vol.~13, 2020.

\bibitem{li2020localization}
Q.~Li, P.~Nevalainen, J.~Pe{\~n}a~Queralta, J.~Heikkonen, and T.~Westerlund,
  ``Localization in unstructured environments: Towards autonomous robots in
  forests with delaunay triangulation,'' \emph{Remote Sensing}, vol.~12, 2020.

\bibitem{ye2019liomap}
H.~Ye, Y.~Chen, and M.~Liu, ``Tightly coupled 3d lidar inertial odometry and
  mapping,'' in \emph{2019 International Conference on Robotics and Automation
  (ICRA)}.\hskip 1em plus 0.5em minus 0.4em\relax IEEE, 2019.

\bibitem{xu2022robust}
Y.~Xu, J.~Lin, J.~Shi, G.~Zhang, X.~Wang, and H.~Li, ``Robust self-supervised
  lidar odometry via representative structure discovery and 3d inherent error
  modeling,'' \emph{IEEE Robotics and Automation Letters}, 2022.

\bibitem{tampuu2022lidar}
A.~Tampuu, R.~Aidla, J.~A. van Gent, and T.~Matiisen, ``Lidar-as-camera for
  end-to-end driving,'' \emph{arXiv preprint arXiv:2206.15170}, 2022.

\bibitem{xianjia2022analyzing}
Y.~Xianjia, S.~Salimpour, J.~P. Queralta, and T.~Westerlund, ``Analyzing
  general-purpose deep-learning detection and segmentation models with images
  from a lidar as a camera sensor,'' \emph{arXiv preprint arXiv:2203.04064},
  2022.

\bibitem{lin2020loam}
J.~Lin and F.~Zhang, ``Loam livox: A fast, robust, high-precision lidar
  odometry and mapping package for lidars of small fov,'' in \emph{2020 IEEE
  International Conference on Robotics and Automation (ICRA)}.\hskip 1em plus
  0.5em minus 0.4em\relax IEEE, 2020.

\bibitem{li2021towards}
K.~Li, M.~Li, and U.~D. Hanebeck, ``Towards high-performance
  solid-state-lidar-inertial odometry and mapping,'' \emph{IEEE Robotics and
  Automation Letters}, vol.~6, no.~3, 2021.

\bibitem{sier2022benchmark}
H.~Sier, L.~Qingqing, Y.~Xianjia, J.~P. Queralta, Z.~Zou, and T.~Westerlund,
  ``A benchmark for multi-modal lidar slam with ground truth in gnss-denied
  environments,'' \emph{arXiv preprint arXiv:2210.00812}, 2022.

\bibitem{jiao2021robust}
J.~Jiao, H.~Ye, Y.~Zhu, and M.~Liu, ``Robust odometry and mapping for
  multi-lidar systems with online extrinsic calibration,'' \emph{IEEE
  Transactions on Robotics}, 2021.

\bibitem{chen2021backpack}
P.~Chen, W.~Shi, S.~Bao, M.~Wang, W.~Fan, and H.~Xiang, ``Low-drift odometry,
  mapping and ground segmentation using a backpack lidar system,'' \emph{IEEE
  Robotics and Automation Letters}, vol.~6, no.~4, 2021.

\bibitem{qin2018vins}
T.~Qin, P.~Li, and S.~Shen, ``Vins-mono: A robust and versatile monocular
  visual-inertial state estimator,'' \emph{IEEE Transactions on Robotics},
  vol.~34, no.~4, 2018.

\bibitem{li2022dataset}
L.~Qingqing, Y.~Xianjia, J.~P. Queralta, and T.~Westerlund, ``Multi-modal lidar
  dataset for benchmarking general-purpose localization and mapping
  algorithms,'' in \emph{2022 IEEE/RSJ International Conference on Intelligent
  Robots and Systems (IROS)}.\hskip 1em plus 0.5em minus 0.4em\relax IEEE,
  2022.

\bibitem{vizzo2023kiss}
I.~Vizzo, T.~Guadagnino, B.~Mersch, L.~Wiesmann, J.~Behley, and C.~Stachniss,
  ``Kiss-icp: In defense of point-to-point icp simple, accurate, and robust
  registration if done the right way,'' \emph{IEEE Robotics and Automation
  Letters}, 2023.

\bibitem{segal2009gicp}
A.~Segal, D.~Haehnel, and S.~Thrun, ``Generalized-icp.'' in \emph{Robotics:
  science and systems}, 2009.

\bibitem{zhang2014loam}
J.~Zhang and S.~Singh, ``Loam: Lidar odometry and mapping in real-time.'' in
  \emph{Robotics: Science and Systems}, vol.~2, no.~9, 2014.

\bibitem{zhou2021plane}
L.~Zhou, D.~Koppel, and M.~Kaess, ``Lidar slam with plane adjustment for indoor
  environment,'' \emph{IEEE Robotics and Automation Letters}, vol.~6, no.~4,
  2021.

\bibitem{xu2021fast}
W.~Xu and F.~Zhang, ``Fast-lio: A fast, robust lidar-inertial odometry package
  by tightly-coupled iterated kalman filter,'' \emph{IEEE Robotics and
  Automation Letters}, vol.~6, no.~2, 2021.

\bibitem{lin2020decentralized}
J.~Lin, X.~Liu, and F.~Zhang, ``A decentralized framework for simultaneous
  calibration, localization and mapping with multiple lidars,'' in
  \emph{IEEE/RSJ IROS}.\hskip 1em plus 0.5em minus 0.4em\relax IEEE, 2020.

\bibitem{shan2020lio}
T.~Shan, B.~Englot, D.~Meyers, W.~Wang, C.~Ratti, and D.~Rus, ``Lio-sam:
  Tightly-coupled lidar inertial odometry via smoothing and mapping,'' in
  \emph{2020 IEEE/RSJ international conference on intelligent robots and
  systems (IROS)}.\hskip 1em plus 0.5em minus 0.4em\relax IEEE, 2020.

\bibitem{ye2019tightly}
H.~Ye, Y.~Chen, and M.~Liu, ``Tightly coupled 3d lidar inertial odometry and
  mapping,'' in \emph{2019 IEEE International Conference on Robotics and
  Automation (ICRA)}.\hskip 1em plus 0.5em minus 0.4em\relax IEEE, 2019.

\bibitem{shan2021lvi}
T.~Shan, B.~Englot, C.~Ratti, and D.~Rus, ``Lvi-sam: Tightly-coupled
  lidar-visual-inertial odometry via smoothing and mapping,'' in \emph{2021
  IEEE international conference on robotics and automation (ICRA)}.\hskip 1em
  plus 0.5em minus 0.4em\relax IEEE, 2021.

\bibitem{nguyen2021miliom}
T.-M. Nguyen, S.~Yuan, M.~Cao, L.~Yang, T.~H. Nguyen, and L.~Xie, ``Miliom:
  Tightly coupled multi-input lidar-inertia odometry and mapping,'' \emph{IEEE
  Robotics and Automation Letters}, vol.~6, no.~3, 2021.

\bibitem{ceressolver}
\BIBentryALTinterwordspacing
S.~Agarwal and K.~Mierle. Ceres solver. [Online]. Available:
  \url{http://ceres-solver.org}
\BIBentrySTDinterwordspacing

\bibitem{dellaert2012factor}
F.~Dellaert, ``Factor graphs and gtsam: A hands-on introduction,'' Georgia
  Institute of Technology, Tech. Rep., 2012.

\bibitem{lixia2012software}
M.~Lixia, A.~Benigni, A.~Flammini, C.~Muscas, F.~Ponci, and A.~Monti, ``A
  software-only ptp synchronization for power system state estimation with
  pmus,'' \emph{IEEE Transactions on Instrumentation and Measurement}, vol.~61,
  no.~5, 2012.

\bibitem{shan2018lego}
T.~Shan and B.~Englot, ``Lego-loam: Lightweight and ground-optimized lidar
  odometry and mapping on variable terrain,'' in \emph{2018 IEEE/RSJ
  International Conference on Intelligent Robots and Systems (IROS)}.\hskip 1em
  plus 0.5em minus 0.4em\relax IEEE, 2018.

\end{thebibliography}
